\documentclass{article}
\usepackage[margin=1in]{geometry}
\usepackage{cite}
\usepackage[utf8]{inputenc}
\usepackage{amsmath,amssymb} %
\usepackage{xcolor}
\usepackage{float}
\usepackage{algorithm2e}
\usepackage{algorithmic}
\usepackage{subcaption}
\usepackage{csquotes}
\usepackage{url}
\usepackage{array}
\usepackage[thinlines]{easytable}
\usepackage{appendix}
\usepackage{ulem}
\usepackage{hyperref} 
\usepackage{booktabs}
\newcolumntype{L}{>{\centering\arraybackslash}m{4cm}}
\usepackage{lipsum}
\usepackage{graphicx}
\usepackage{caption}
\usepackage{breqn}
\usepackage{array}
\usepackage{geometry}
\graphicspath{{./Figures/}{../Figures}}
\RestyleAlgo{ruled}

\title{Hall Effect Thruster Forecasting using a Topological Approach for Data Assimilation}

\author{
    \textbf{Max M.~Chumley}\\
    chumleym@msu.edu\\
    Michigan State University\\
    \and
    \textbf{Firas A.~Khasawneh$^*$}\\
    khasawn3@msu.edu\\
    Michigan State University\\
}

\date{\today}

\begin{document} 

\maketitle
*Address all correspondence to this author.
\section{Abstract}
Hall Effect Thrusters (HETs) are electric thrusters that eject heavy ionized gas particles from the spacecraft to generate thrust. Although traditionally they were used for station keeping, recently They have been used for interplanetary space missions due to their high delta-V potential and their operational longevity in contrast to other thrusters, e.g., chemical. However, the operation of HETs involves complex processes such as ionization of gases, strong magnetic fields, and complicated solar panel power supply interactions. Therefore, their operation is extremely difficult to model thus necessitating Data Assimilation (DA) approaches for estimating and predicting their operational states. Because HET's operating environment is often noisy with non-Gaussian sources, this significantly limits applicable DA tools.
We describe a topological approach for data assimilation that bypasses these limitations that does not depend on the noise model, and utilize it to forecast spatiotemporal plume field states of HETs. Our approach is a generalization of the Topological Approach for Data Assimilation (TADA) method that allows including different forecast functions. We show how TADA can be combined with the Long Short-Term Memory network for accurate forecasting. We then apply our approach to high-fidelity Hall Effect Thruster (HET) simulation data from the Air Force Research Laboratory (AFRL) rocket propulsion division where we demonstrate the forecast resiliency of TADA on noise contaminated, high-dimensional data.

\section{Introduction}

\begin{figure}[htbp]
    \begin{center} 
      \includegraphics[width=0.22\textwidth]{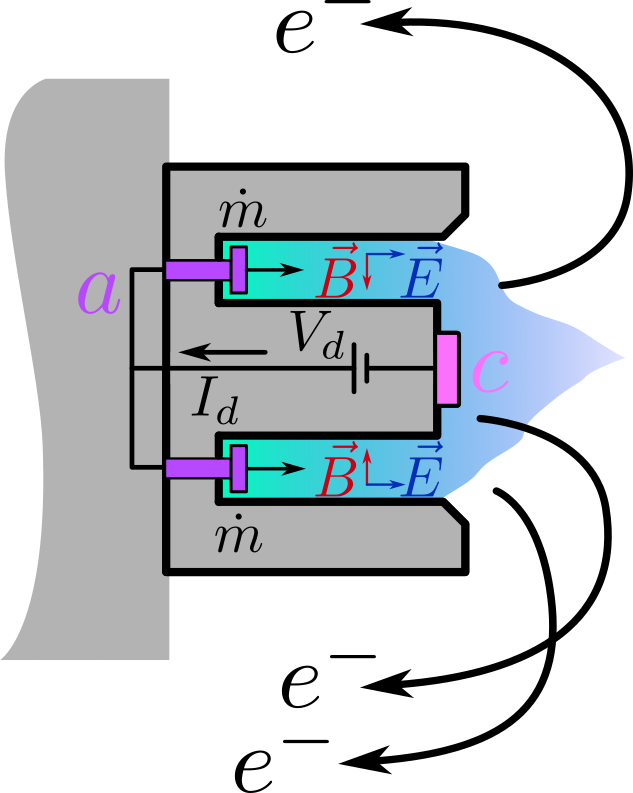}
    \end{center}
     \caption{Hall-Effect 
  Thruster (HET).}
     \label{fig:het_schematic}
  \end{figure}

HETs are a class of ion thrusters that generate thrust by accelerating ions through an electromagnetic field to eject heavy ionized gas particles from the spacecraft. These thrusters are highly complex due to the interplay of electrodynamics, fluid dynamics, fluid-structure interaction, and quantum mechanics which makes them difficult or impossible to model analytically. This is further complicated in experimental settings where the thruster is placed in a vacuum chamber to simulate the environment of space. The electromagnetic field interacts with the chamber in these experiments leaving researchers with data that likely does not accurately represent the thrusters behavior in space due to ground effects. HETs have shown great potential for future space flight due to their ability to greatly increase the lifespan of the thruster to over 10,000 hours \cite{Bapat2022}. However, some of the operating modes in the HETs lead to undesirable system dynamics such as high amplitude, low frequency breathing mode oscillations in the thrust produced. This phenomena is due to a complex interaction between neutral and ionized particles and leads to sub-optimal performance of the thruster \cite{Hara2019}. Another behavior that these thrusters exhibit is a result of high energy ions causing erosion of critical surfaces for the thruster and the space craft which is detrimental to many of the components on board \cite{Hara2019}. 
These operating modes are induced by changes to the system parameters (shown in Fig.~\ref{fig:het_schematic}) such as the discharge voltage $V_d$ (the voltage between the anode $a$ and cathode $c$), mass flow rate of gas $\dot{m}$, magnetic field strength and topology $\vec{B}$, electric field strength $\vec{E}$ and discharge current $I_d$ \cite{Hara2019, Bapat2022}. Thus the ability to develop accurate and optimal data-driven models to predict future system states based on measurement data is crucial for safe operation of these thrusters. This makes HETs a good candidate for time series forecasting and data assimilation methods such as our recently developed algorithm Topological Approach for Data Assimilation (TADA) \cite{chumley2024topological}.

\begin{figure}[htbp]
  \begin{minipage}[b]{0.5\linewidth}
    \centering
    \includegraphics[width=\textwidth]{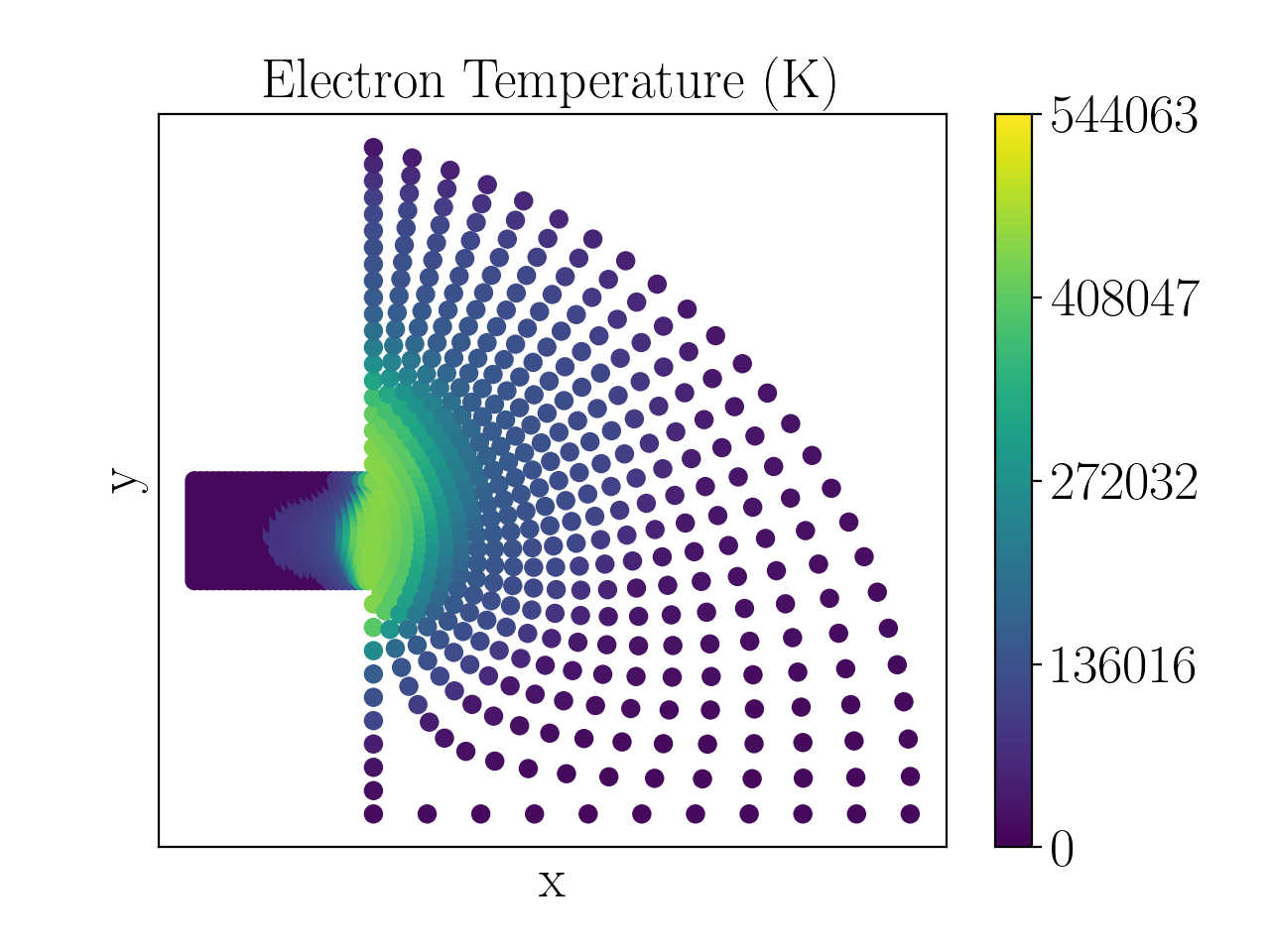}
    (a) Data Locations
  \end{minipage}
  \begin{minipage}[b]{0.5\linewidth}
    \centering
    \includegraphics[width=\textwidth]{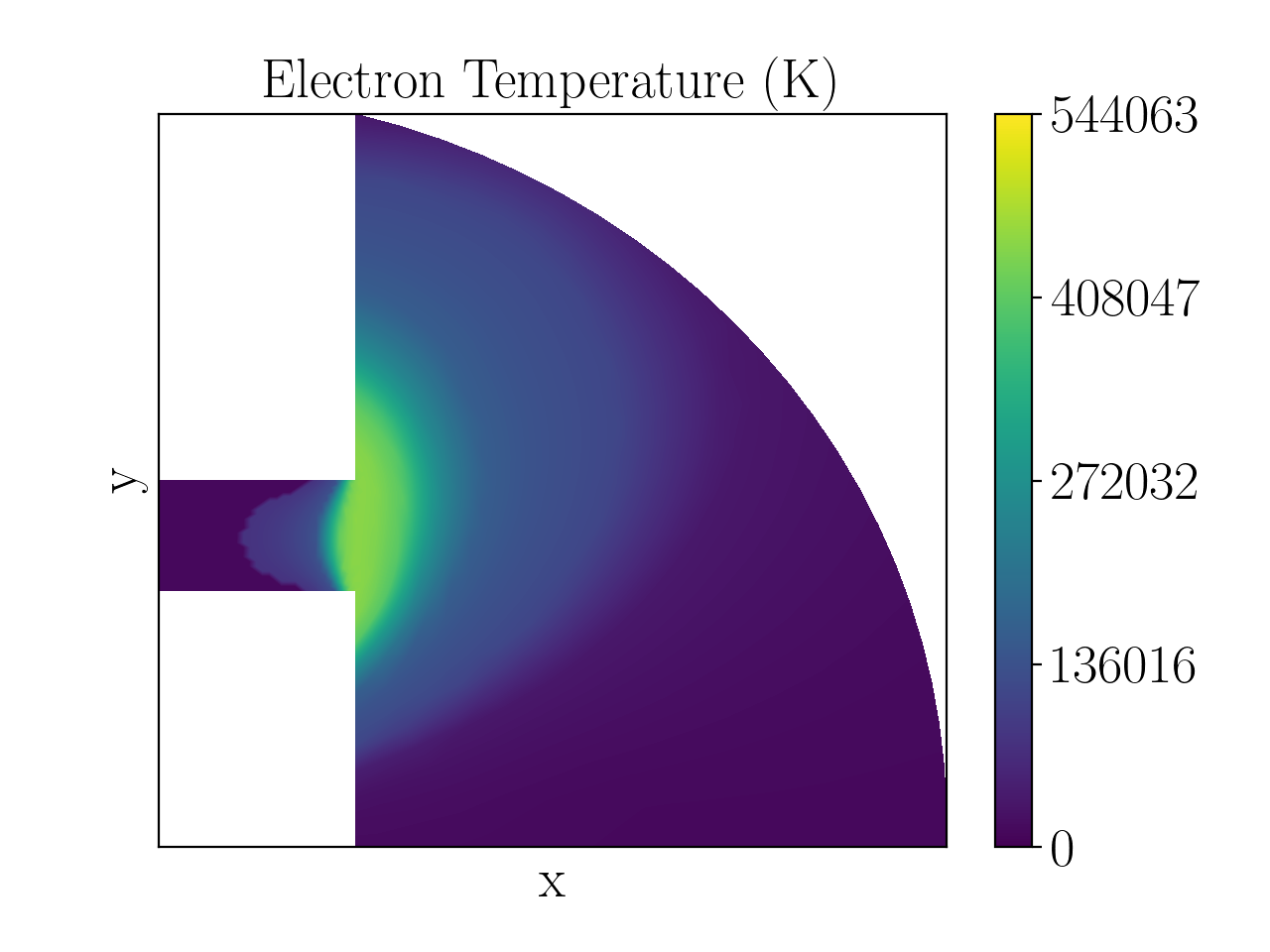}
    (b) Contour Plot
  \end{minipage}
  \caption{Example plots of HET data using a single frame of the electron temperature signals. (a) shows a scatter plot of the spatial locations at one moment in time and (b) shows the corresponding contour plot of the field.}
  \label{fig:thruster_plume}
\end{figure}

\begin{figure}[htbp]
  \centering
  \includegraphics[width=0.8\textwidth]{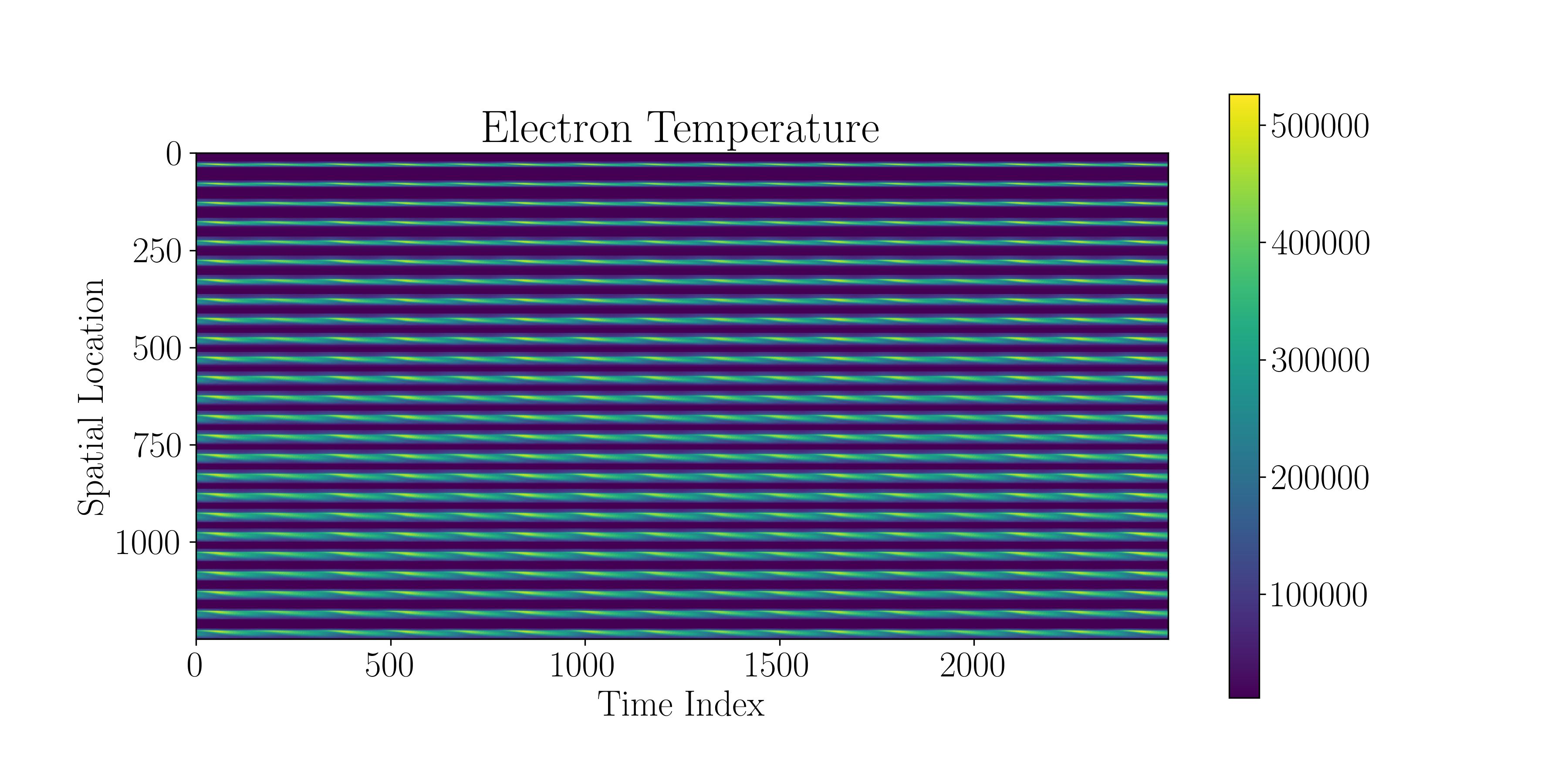}
  \caption{Electron temperature data matrix for the first 2,500 time points.}
  \label{fig:data_matrix}
\end{figure}

  The data from AFRL was generated using a software package called HPHall which uses a hybrid Particle In Cell (PIC) method to model particles on different scales in the system \cite{Fife1998} and the Direct Simulation Monte Carlo (DSMC) method for simulating collisions \cite{Smith_2013}. Specifically, the SPT-100 thruster was simulated for this work at a mass flow rate of $\dot{m}=5.01\times 10^{-6}~kg/s$ at a discharge voltage $V_d=300~V$. While this data is obtained from a simulation, the highly complex behavior of these thrusters make them incredibly difficult to accurately model analytically, and the methods used for simulation ultimately leave the data contaminated with noise with unknown statistics. Simulation data was provided over 20,000 time steps at 1,250 locations in the thruster plume measuring 7 field states at each positional location. The measured states include electron temperature $T_e$, electric potential $\phi$, neutral number density $n_n$, electron number density $n_e$, ion production rate $\dot{n}_i$, axial ion velocity $v_{iz}$ and radial ion velocity $v_{ir}$. Due to the radial symmetry of the thruster, the simulation data forms a radial slice in the thruster plume. Figure~\ref{fig:thruster_plume} shows plots of the thruster plume at a single point in time using the electron temperature field. 
We see that the grid of points is highly nonuniform due to the optimizations performed by HPHall. When using the provided spatial locations for each point, the data can be arranged into this form to view the behavior of the thrust field, but in practice it is easier to work with a data matrix $X\in \mathrm{R}^{n_x\times n_t}$ where $n_x=1,250$ is the number of spatial locations and $n_t=20,000$ is the number of time points. Slices of this data matrix can also be visualized as images to study spatio-temporal patterns. Figure~\ref{fig:data_matrix} shows the transpose of the data matrix for the first 2,500 time points for the electron temperature. 
The figure shows an oscillating pattern present in some spatial locations while other locations show constant temperature. The latter points largely correspond to the lower right corner of Fig.~\ref{fig:thruster_plume}.

Using this data matrix, we aim to use TADA to train forecasting models to predict future states of the system and perform optimal updates to the model as new measurements become available. This paper is structured as follows. We present the relevant background in Section~\ref{sec:methods} including persistent homology and persistence optimization. This section also includes the theoretical framework for the forecasting method we use, Long Short-Term Memory networks (LSTM). In Section~\ref{sec:theory}, we generalize the TADA algorithm to include other forecast functions and demonstrate that the LSTM network fits this framework. Section~\ref{sec:results} includes TADA results for two HET field variables while the conclusions are in Section~\ref{sec:conclusion} .

\section{Background} \label{sec:methods}
In this section, we provide the required background for understanding the TADA algorithm. The first tool comes from Topological Data Analysis (TDA) and is called persistent homology. 

\subsection{Persistent Homology}
\begin{figure}
    \centering
    \includegraphics[width=\textwidth]{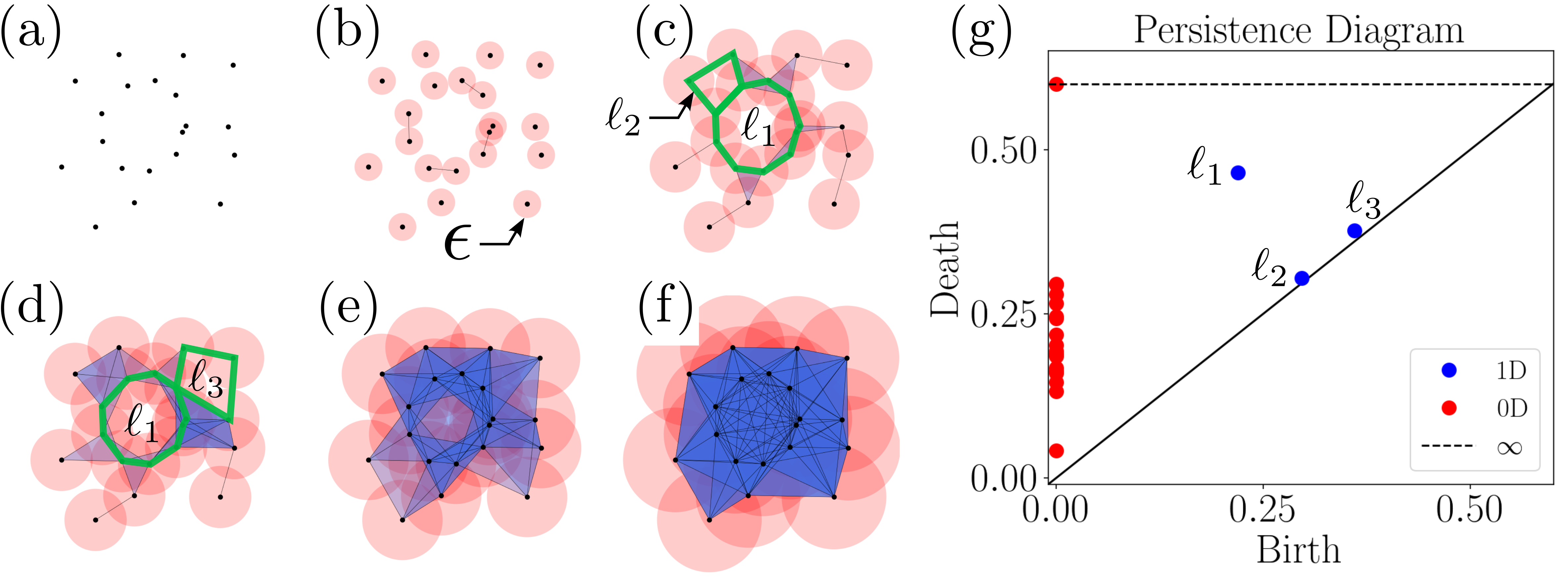}
    \caption{Example of point cloud persistence using the Vietoris-Rips complex. The frames in (a)--(e) show the simplicial complex at various stages of the filtration and values of the connectivity parameter $\epsilon$. The point cloud contains three loops, one of which is prominent ($\ell_1$) and persists relatively far from the diagonal with two smaller loops close to the diagonal ($\ell_2$ and $\ell_3$). These loops are shown in the 1D persistence diagram in (g) as (birth,death) pairs. The 0D persistence pairs (connected components) are also shown as red points along with the infinite persistence pair representing the fully connected component.}
    \label{fig:persistence_example}
\end{figure}

Topological Data Analysis is comprised of a set of tools for quantifying shape information from various types of data. The main form of data we are focused on for this paper is point cloud data where we have a set of points in $\mathrm{R}^n$ and we wish to analyze the shape of the point cloud to draw conclusions about the system that produced the data. For this application, we typically use point cloud persistent homology. Persistent homology has many benefits such as its stability under small perturbations \cite{CohenSteiner2006}, and its ability to provide a compact representation of complex structures in data. This works by inducing a simplicial complex on the point cloud which for this paper is the Vietoris-Rips complex. A simplicial complex can be thought of as a generalized graph where instead of only having vertices and edges we can also have faces and higher dimensional simplices like tetrahedra. We can study the shape or structure of a simplicial complex by computing its homology in different dimensions. For example, the rank of the 0D homology of a simplicial complex is the number of connected components and for 1D this is the number of loops and so on for higher dimensions. With persistent homology, we parametrize the simplicial complex based on a connectivity parameter $\epsilon$ which for the Vietoris-Rips complex represents the radius of the balls centered at each point. As $\epsilon$ increases, we allow the simplicial complex to change based on the distances between points. Any two points with a distance of $2\epsilon$ or less are connected with an edge and any three points with this property are connected with a face or triangle. This process yields a filtration where the previous complex is always a subset of the next one. The homology is computed for all values of $\epsilon$ and the birth and death of topological features are tracked as the horizontal and vertical coordinates in a persistence diagram. An example of this process is shown in Fig.~\ref{fig:persistence_example} where we see  a simple point cloud consisting of two concentric circles with additive noise. These points can be thought of as the state space point cloud of a dynamical system. As $\epsilon$ increases, we see the radius of the balls increasing with edges and faces being added. We see in Fig.~\ref{fig:persistence_example}(a), all 20 components are born with $\epsilon=0$ and the components die when they connect to an older component. These 0D persistence pairs are plotted in red in Fig.~\ref{fig:persistence_example}(g). We also see the infinite persistence pair plotted on the dashed line representing the final connected component that persists forever in Fig.~\ref{fig:persistence_example}(f). For 1D persistence, we see in Fig.~\ref{fig:persistence_example}(c) that two small loops are present and outlined in green ($\ell_1$ and $\ell_2$). $\ell_2$ is a small loop and we see the corresponding persistence pair is close to the diagonal indicating that it is likely not a prominent feature of the data. $\ell_1$ however, is a larger loop and its persistence pair is much further from the diagonal so the persistence diagram encodes this shape information. Likewise to $\ell_2$, $\ell_3$ is a small loop shown in Fig.~\ref{fig:persistence_example}(d) and it quickly dies indicating it is not significant to the overall shape of the data. For more information on TDA, we refer the interested reader to \cite{Hatcher2002,Kaczynski2004,Ghrist2008,Carlsson2009,Edelsbrunner2010,Mischaikow2013,oudot2017persistence,Munch2017}.

\subsection{Persistence Optimization}
\begin{figure}
    \centering
    \includegraphics[width=\textwidth]{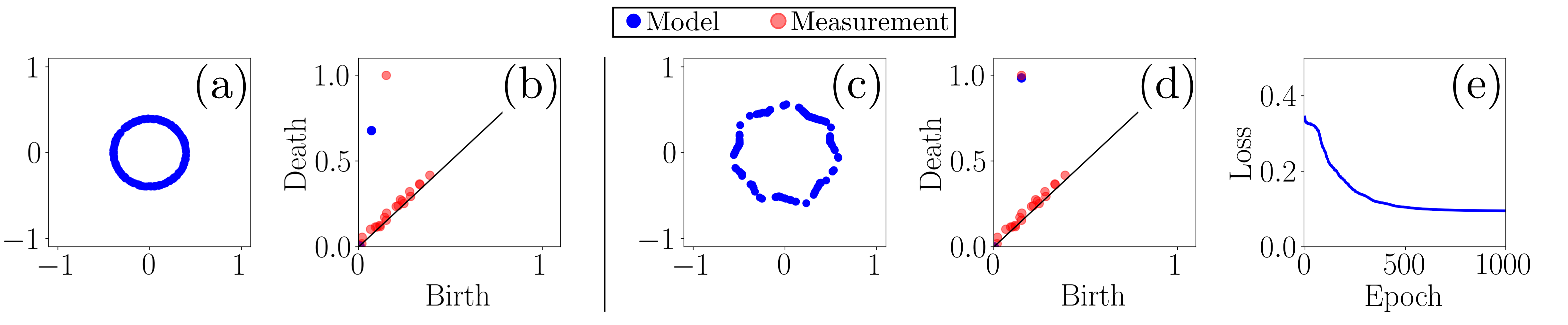}
    \caption{Example minimizing the Wasserstein distance persistence function to reach a point cloud with a target persistence diagram. (a) shows the original point cloud and (b) shows the original persistence diagram and target persistence diagram. (c) shows the optimized point cloud and (d) shows the optimized persistence diagram with the loss plot in (e). }
    \label{fig:wd_minimize}
\end{figure}

Persistence diagrams are often converted to real-valued features using functions of persistence for quantifying various topological properties of the data. For example we can compute the total persistence of a 1D persistence diagram using $F(PD_1)=\sum_{i=1}^p |d_i-b_i|$, where $PD_1$ is the 1D persistence diagram, $b_i$ is the $i$th birth value and $d_i$ is the $i$th death value. $F$ is the sum of all persistence lifetimes in the diagram and provides a measure of how large the loops are in the data. Recent advancements in TDA have enabled differentiability of persistence diagrams for optimizing functions of persistence using gradient descent \cite{Carriere2020, Leygonie2021, Gameiro2016}. Specific details on how persistence diagrams are differentiated are shown in our original paper introducing the TADA algorithm \cite{chumley2024topological}. We use \texttt{tensorflow} and its automatic differentiation framework along with the \texttt{gudhi} Python library \cite{Carriere2020} for topological data analysis to minimize persistence based loss functions using gradient descent. Specifically for TADA, our loss function is based on the Wasserstein distance which gives a measure of dissimilarity for two persistence diagrams. This function can be minimized to find a point cloud with a persistence diagram that matches a given target persistence diagram. We show an example of this process in Fig.~\ref{fig:wd_minimize} where we start with a simple circular point cloud in Fig.~\ref{fig:wd_minimize}(a) and the blue persistence diagram in Fig.~\ref{fig:wd_minimize}(b) and the goal is to reach a point cloud with the target persistence diagram or red points in Fig.~\ref{fig:wd_minimize}(b). This persistence diagram has a pair that is further from the diagonal but also consists of many pairs near the diagonal to simulate noise. 
Using persistence optimization, we minimize the Wasserstein distance and reach the point cloud in Fig.~\ref{fig:wd_minimize}(c) with persistence diagram shown in Fig.~\ref{fig:wd_minimize}(d). It is clear that the new point cloud picked up the prominent persistence feature while avoiding the new features due to noise because the Wasserstein distance function in \texttt{gudhi} does not add features to the persistence diagrams. The loss landscape is also shown in Fig.~\ref{fig:wd_minimize}(e) where we see a clear minimum was reached, but not at zero due to the noise persistence pairs.

\subsection{Long Short-Term Memory (LSTM) Networks}
Now we transition away from TDA to show the forecast function that we use in this paper. The Long Short-Term Memory (LSTM) network is an advanced Recurrent Neural Network (RNN) approach for time series forecasting. LSTM networks are composed of LSTM units rather than neurons like a traditional Feed Forward Neural Network (FNN) or RNN. Over time with RNNs, the backpropagation gradients will either explode or vanish due to the low memory bandwidth of these networks (generally 5--10 time steps) \cite{Staudemeyer2019}. If the gradient explodes, this leads to extreme oscillations in the model weights that do not converge to give good predictions and if the gradient vanishes the model essentially stops learning \cite{Staudemeyer2019}. LSTM aims to mitigate these issues by incorporating memory cells, input gates, output gates, and forget gates into the network architecture \cite{Staudemeyer2019}. The inputs to the cell are split into long term and short term memories that consist of input signal states \cite{geron2022hands}. The long term memories are represented as $c_t$ and short term memories are $h_t$. These memories get passed through a \textit{forget layer} ($f_t$) which uses a sigmoid activation function $\sigma$ to decide which memories to drop from $c_t$ . This is represented mathematically as
\begin{equation}\label{eq:forget_layer}
    f_t = \sigma(W_{xf}^T x_t + W_{hf}^T h_{t-1} + b_f),
\end{equation}
where $W_{xf}$ and $W_{hf}$ are weight matrices corresponding to the current input $x_t$ and short term memory at the previous step $h_{t-1}$ and $b_f$ is a bias \cite{geron2022hands}. Next, the \textit{main layer} $g_t$ which takes the current state and short term memory states and acts as a traditional RNN activation layer using corresponding weight matrices and bias vector with the activation equation \cite{geron2022hands}
\begin{equation}\label{eq:lstm_main_layer}
    g_t = \tanh{(W_{xg}^T x_t + W_{hg}^T h_{t-1} + b_g)}.
\end{equation}
the \textit{input layer} is then used to determine which memories are added to the long term memory $c_t$ using the sigmoid function \cite{geron2022hands}
\begin{equation}\label{eq:input_gate}
    i_t = \sigma(W_{xi}^T x_t + W_{hi}^T h_{t-1} + b_i),
\end{equation}
with similar weights and biases. The long term memories $c_t$ are updated by incorporating the output from the $o_t$ and $i_t$ using \cite{geron2022hands}
\begin{equation}\label{eq:lstm_long_term_update}
    c_t = f_t \odot c_{t-1} + i_t \odot g_t,
\end{equation}
where $\odot$ is element-wise multiplication. Equation~\eqref{eq:lstm_long_term_update} serves as the sum of the \textit{forget gate} and \textit{input gate} \cite{geron2022hands}. Lastly, the short term memories are updated using an \textit{output gate} to determine which part of the long term memories should be the output of the cell using \cite{geron2022hands}
\begin{equation}\label{eq:output_gate}
    o_t = \sigma(W_{xo}^T x_t + W_{ho}^T h_{t-1} + b_o),
\end{equation}
and the short term memories or predictions of the next signal states are then given by the output gate equation \cite{geron2022hands},
\begin{equation}\label{eq:lstm_output}
    h_t = o_t \odot \tanh{(c_t)}.
\end{equation}
Therefore, using Equations~\eqref{eq:forget_layer} -- \eqref{eq:lstm_output} backpropagation can be used to update the model weights and biases using training data to provide future predictions for the states by passing in sequences of past training points with the outputs being the corresponding next points in the sequence. 

\section{Theory}\label{sec:theory}

In this section, the TADA algorithm is generalized to include a general class of forecast functions that utilize sequences of past points to predict future states and we integrate training data into the algorithm to increase its robustness. We also demonstrate that the LSTM forecast function fits this general framework.

\subsection{Topological Approach for Data Assimilation}
Here we aim to generalize the framework for the TADA \cite{chumley2024topological} algorithm.  This algorithm leverages persistence optimization to optimally combine measurement data and model predictions. In \cite{chumley2024topological}, we introduce a special case of the algorithm using reservoir computing for generating the data driven models for the system. In this paper, we generalize TADA to fit other forecast models and introduce features to improve its reliability. 

The TADA algorithm assumes that we have access to $N$ system state measurements that are contaminated with noise. A subset of measurements are then used for training a data driven model that fits the form of Eq.~\eqref{eq:general_forecast} and is differentiable with respect to the model weights. The forecast model can be defined as,
\begin{equation}\label{eq:general_forecast}
    \mathbf{x}_{n+1}=G(\mathbf{X}_p, \mathbf{w}, \mu),
\end{equation}
where $\mathbf{x}\in\mathrm{R}^N$ is a vector of system states, $n$ is the time index, $\mathbf{X}_p=(\mathbf{x}_{n-p}, \dots, \mathbf{x}_{n-1},\mathbf{x}_{n})$ is a matrix of the $p+1$ previous states, $w$ is the set of model weights that determine the output of the forecast function $G$ and $\mu$ is a set of hyperparameters such as the hidden states in the LSTM network. In general, the training data is used to minimize a cost function that quantifies the error between the measurements and predictions by tuning $\mathbf{w}$. Once a framework for $G$ is fixed, the model is used to forecast the next $\mathcal{W}$ points and measurements are also collected over this window making up the $n$th assimilation window $\mathcal{W}_n$. Each assimilation window gives us two point clouds of length $\mathcal{W}_n$, one for measurements and one for predictions and persistence diagrams are obtained for the point clouds using the Vietoris-Rips filtration. The topological differences between these two point clouds are minimized by minimizing the Wasserstein distance between the persistence diagrams and the model weights $\mathbf{w}_n$ are updated using gradient descent. We take the cost function to be the sum of the 0D and 1D Wasserstein distances for the point cloud persistence diagrams and also include regularization Wasserstein distance terms to reduce temporal shifting \cite{chumley2024topological}. The updated model weights $\mathbf{w}_{n+1}$ are then used for the next assimilation window and the process is continued as new measurements are collected. In \cite{chumley2024topological}, we also sample points from the training set to promote a solution that is close to the original model and the same approach is taken in this work. 

\begin{figure}[htbp]
  \centering 
    \includegraphics[width=0.95\textwidth]{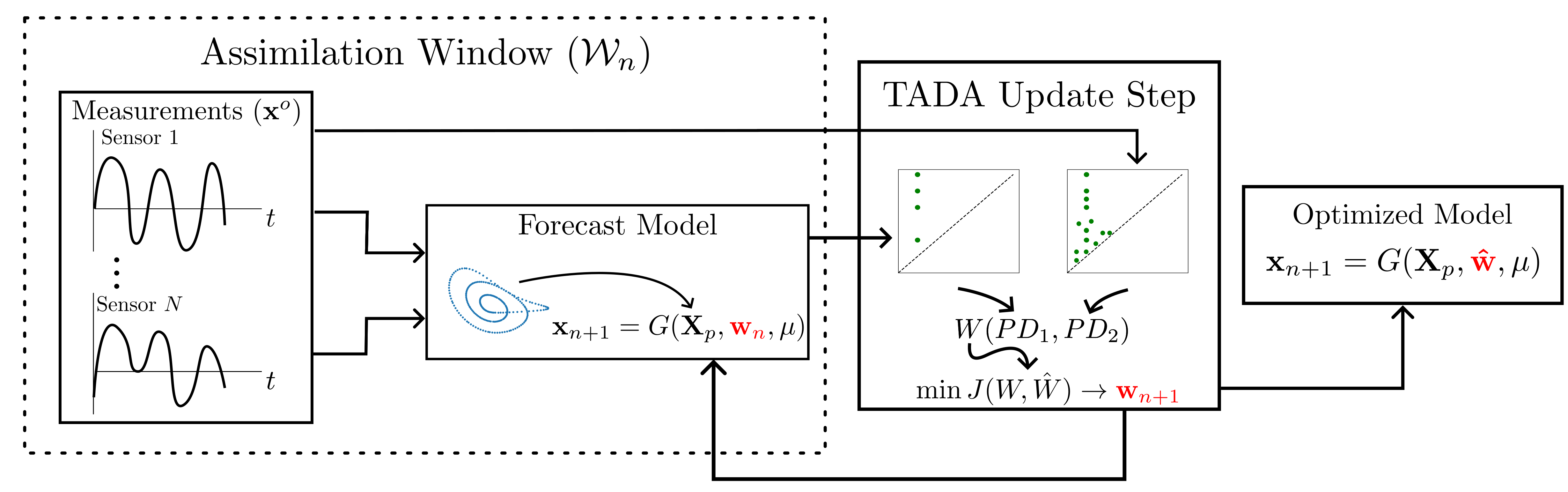}
    \caption{Assimilation window update diagram showing the updated model giving an improved forecast. This algorithm is originally published in \cite{chumley2024topological} and this figure is updated for the generalized version.}
    \label{fig:tada_algorithm}
\end{figure}
The main benefit of TADA is that it does not impose any assumptions on the noise distribution like other data assimilation algorithms so if the noise distribution is unknown this method still allows for optimally updating the model and performing data assimilation.

\subsection{LSTM Forecast Function}

For this application, we chose to use the LSTM forecast function due to its ability to maintain long term memories and they can be trained faster than traditional RNN methods \cite{geron2022hands}. The LSTM framework also allows for inputs and outputs to be sequences of points rather than just a single point which provides a more reliable forecast because more information is contained in the sequence. The LSTM forecast function can be written in the form $\mathbf{x}_{n+1}=G(\mathbf{X}_p, \mathbf{w}, \mu)$ by combining Eqs.~\eqref{eq:forget_layer}--\eqref{eq:lstm_output} with $c_0=0$ and $h_0=0$ to get the following update relationships for the cell and hidden states of the network,
\begin{equation}
\begin{aligned}
  c_n =  &\sigma(W_{xf}^T \mathbf{X}_p + W_{hf}^T h_{n-1} + b_f) \odot c_{n-1} \\
  + &\sigma(W_{xi}^T \mathbf{X}_p + W_{hi}^T h_{n-1} + b_i) \odot \tanh(W_{xg}^T \mathbf{X}_p + W_{hg}^T h_{n-1} + b_g)\\
  h_{n} = &\sigma(W_{xo}^T \mathbf{X}_p + W_{ho}^T h_{n-1} + b_o) \odot \tanh{(c_n)}.
\end{aligned}
\end{equation}
These relationships are then combined with a dense or fully connected layer to map back to the dimension of the input matrix $\mathbf{X}_p$ to predict the next point using the equation,
\begin{equation}
    \mathbf{x}_{n+1} = W_d h_n + b_d,
\end{equation}
where $W_d$ and $b_d$ are the corresponding learned weight matrix and bias vector for the dense layer to predict the next state of the system. These three update rules combined fit the form $\mathbf{x}_{n+1}=G(\mathbf{X}_p, \mathbf{w}, \mu)$ for use with the TADA algorithm where all of the model weights and biases are contained in $\mathbf{w}$ and the initial cell and hidden states are captured in $\mu$. Note that there is not an easily expressed explicit form for $G$ because the update rule for $c_n$ is recursive. Using backpropagation, the model parameters are trained by minimizing the error between predictions and the training data and an optimal model is learned. Note that for this work the inputs are fixed to being the previous 5 states and the output is only a single point for consistency with Eq.~\eqref{eq:general_forecast}. Many other forecast functions also fit this framework such as the random feature map or reservoir computing method used in \cite{chumley2024topological}, the AutoRegressive (AR), Moving Average (MA) and AutoRegressive Integrated Moving Average (ARIMA) models \cite{zhang2003time,mouraud2017innovative}, but for this work we focus on the LSTM method due to the significant improvements it provides in forecast ability over the ARIMA method \cite{siami2018comparison}.

\section{Results}\label{sec:results}

LSTM networks were trained using 5000 HET measurement points and 500 LSTM units were used to generate weights for the system. The models were trained on sequences of 5 points which were used to predict the next point. To demonstrate the robustness of TADA on this data, the models were fit using only 5 optimization epochs with batches of size 50 and the full TADA cost function ($J=J_1+J_2$) was used to improve the model. A window size of 200 points was used for these results to capture the topology of the signals. These results demonstrate successful forecast improvements on high dimensional spatiotemporal data with the ability to predict future points based on previous data. Using the full TADA cost function shows that the model predictions do not necessarily need to be close to the measurements in this case. In this section, the forecast accuracy is quantified using the squared differences between the measurements and model predictions and the color scale on the error plots is set based on the largest difference in the first TADA step. This is because many of the states remain very close or equal to zero so dividing by the measurements to compute a percent error leads to skewed error results. Many machine learning models perform better on data that is scaled or normalized so for these results, all signals were z-normalized. 

\subsection{Electron Temperature ($T_e$)}

\begin{figure}[htbp]
  \begin{minipage}[b]{0.5\linewidth}
    \centering
    \includegraphics[width=\textwidth]{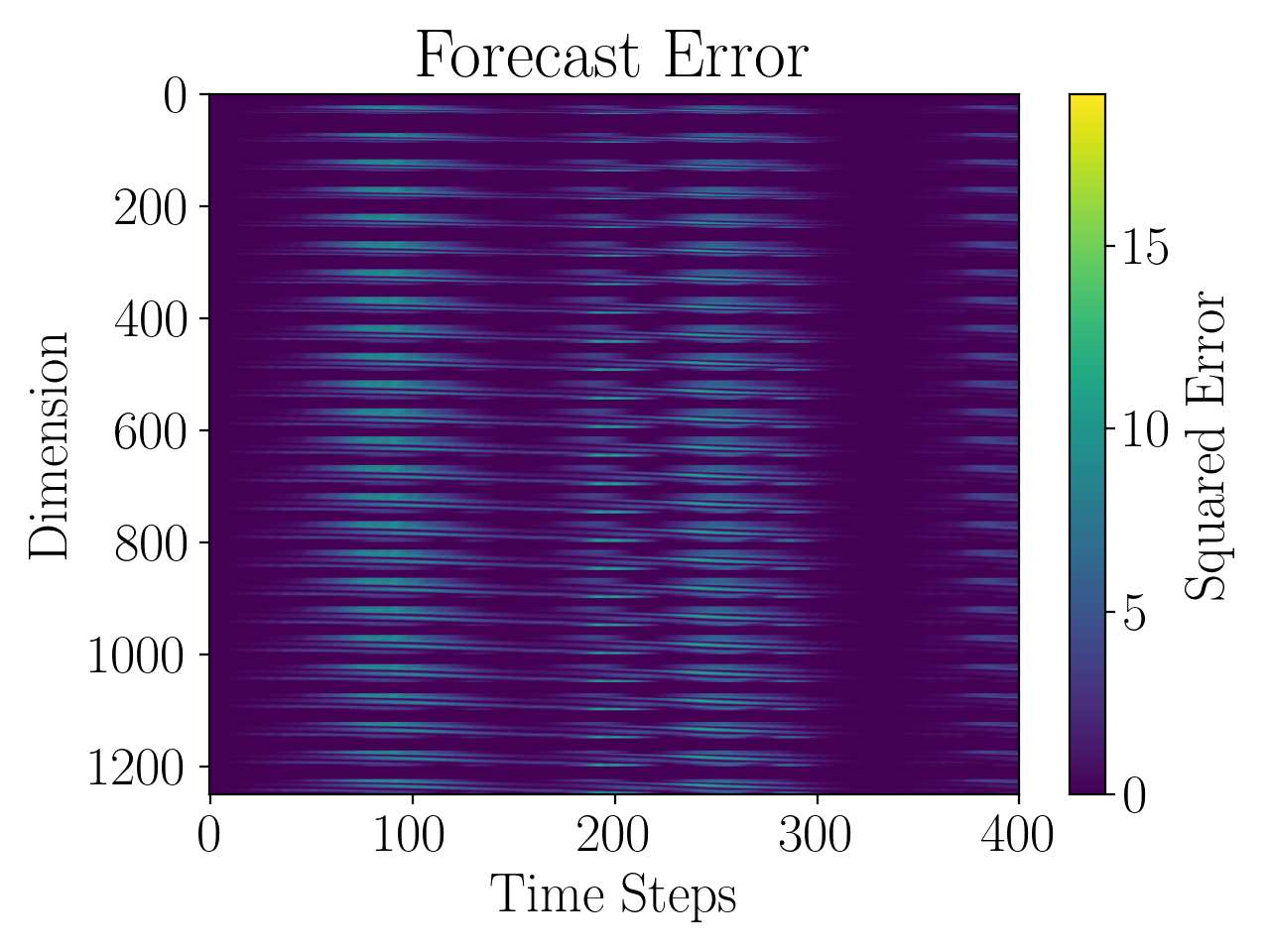}
    (a) Initial $T_e$ error
  \end{minipage}
  \begin{minipage}[b]{0.5\linewidth}
    \centering
    \includegraphics[width=\textwidth]{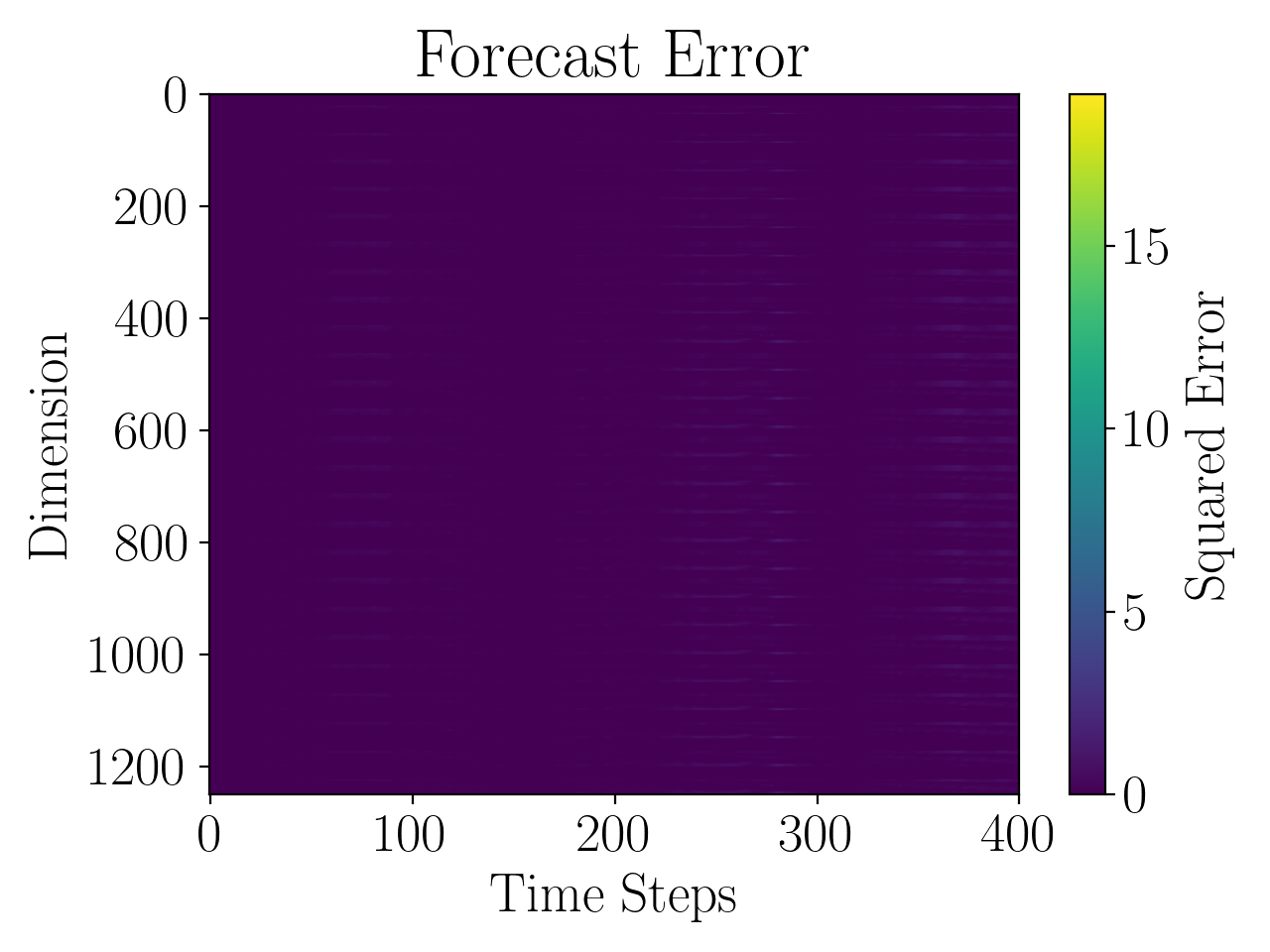}
    (b) Final $T_e$ error
  \end{minipage}
  \caption{Error plots for the electron temperature before and after 200 TADA steps.}
  \label{fig:t_e_error}
\end{figure}

\begin{figure}[htbp]
  \centering
  \includegraphics[width=0.75\textwidth]{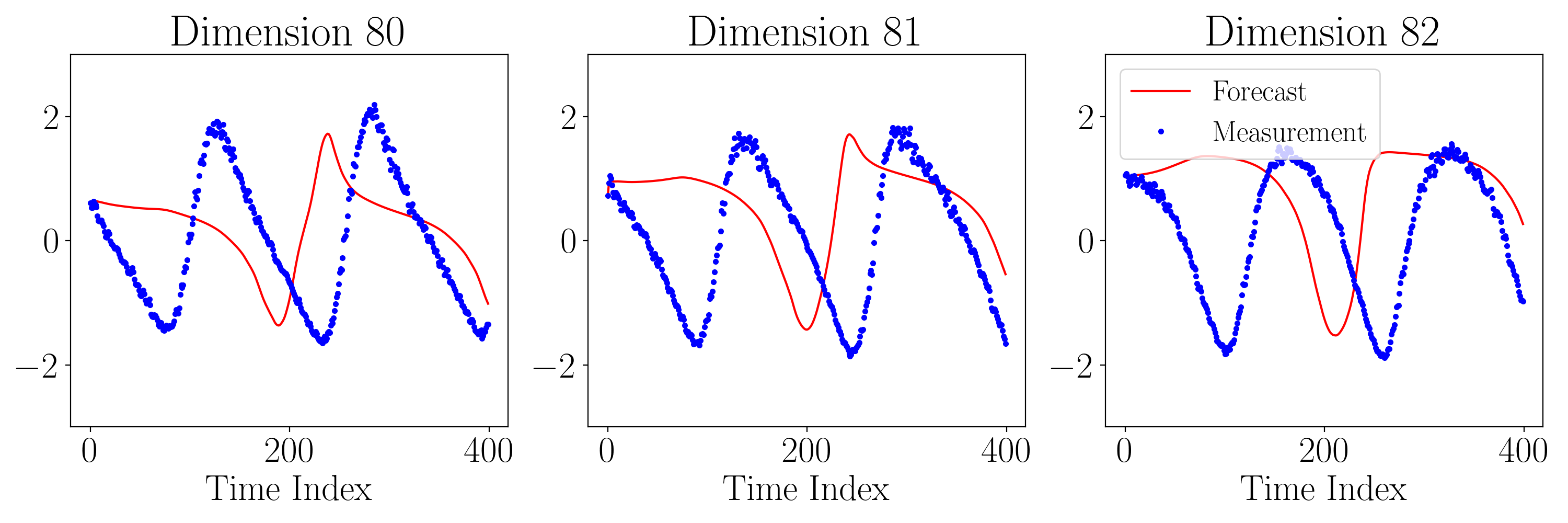}
  \caption{Initial $T_e$ forecast for dimensions 80-82.}
  \label{fig:t_e_initial_forecast}
\end{figure}

The electron temperature ($T_e$) was tested first using this approach. Figures~\ref{fig:t_e_error}(a) and \ref{fig:t_e_initial_forecast} show the initial forecast results for the electron temperature. It is clear from these results that the forecast does not accurately reflect the measurements and the model started quite far from the measurements. TADA was then applied to this model using a learning rate of $10^{-5}$ with a decay of 1\% for every step. Because the $J_2$ cost function term was used, the learning rate decision is not as critical. If the learning rate is too high the predictions oscillate around the measurements eventually converging as the learning rate decays. For this application the learning rate was manually tuned to $10^{-5}$ as this was found to converge quickly while minimizing oscillations of the forecast around the true minimum of the loss function. 

\begin{figure}[htbp]
  \centering
  \includegraphics[width=0.75\textwidth]{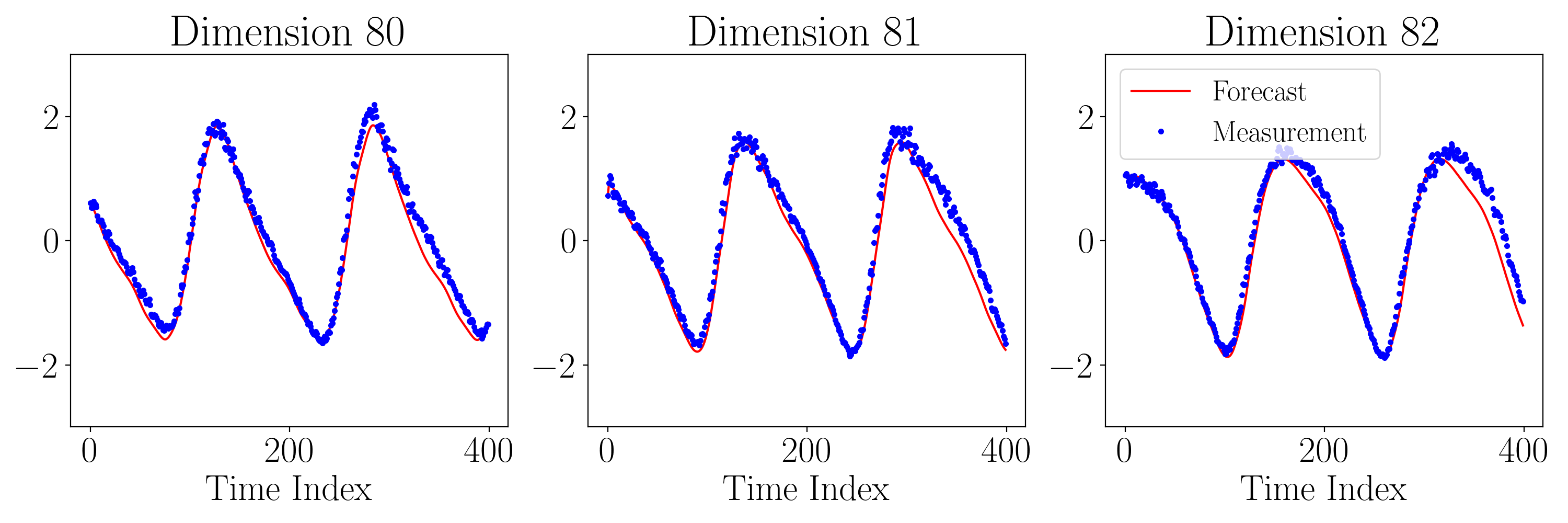}
  \caption{$T_e$ forecast for dimensions 80-82 after 200 TADA steps.}
  \label{fig:t_e_200_forecast}
\end{figure}

After 200 TADA steps, the resulting forecast error and forecast for dimensions 80--82 are shown in Figs.~\ref{fig:t_e_error}(b) and \ref{fig:t_e_200_forecast}. We see that the forecast error decreased dramatically with a maximum initial error of approximately 20 and the final maximum error was 0.98. The forecast shown in Fig.~\ref{fig:t_e_200_forecast} also shows a significantly more accurate prediction that more closely matches the measurements. While 200 points were used for data assimilation, the forecast remains accurate outside of the DA window through the next 200 time points without over fitting to the noise in the signal. 
 
\subsection{Electric Potential ($\phi$)}

\begin{figure}[htbp]
  \centering
    \centering
    \includegraphics[width=\textwidth]{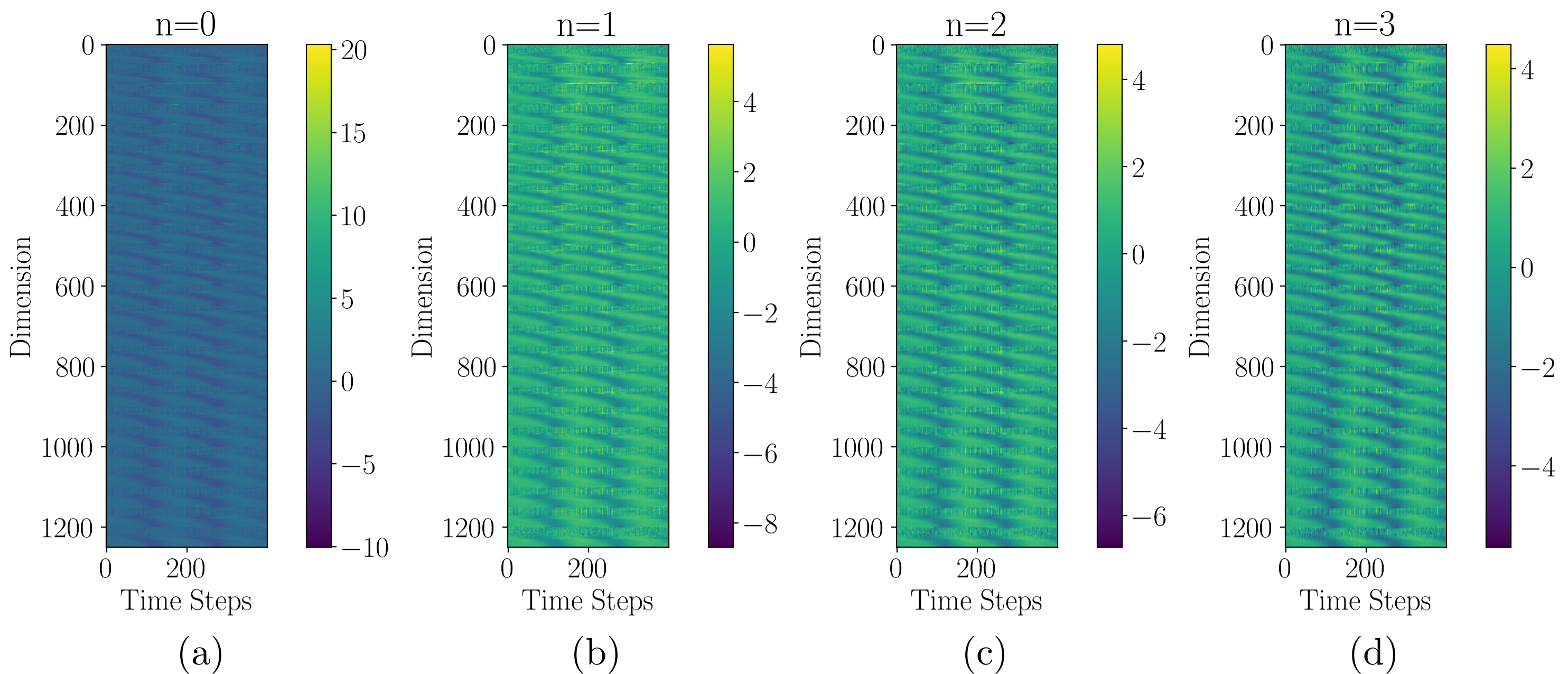}
  \caption{Scaled $\phi$ plots with different levels of smoothing. The plot in (a) contains no smoothing, (b) was generated using $n=1$ smoothing, (c) $n=2$ and (d) $n=3$.}
  \label{fig:scaled_phi_plots}
\end{figure}

The electron temperature data was analyzed first because it is the most well behaved variable in this data set. By well behaved we mean it contained minimal outliers. This is not the case for other HET variables such as the electric potential. If we naively generate LSTM models from the $\phi$ field data, the model does not converge to an accurate solution and running TADA on this model does not improve the forecast due to some of the states containing outliers. This can be easily visualized by plotting the scaled $\phi$ fields. We see in Fig.~\ref{fig:scaled_phi_plots}(a) that the original scaled data with no smoothing appears to have most z-scores relatively close to zero, but some points in the data are as much as 20 standard deviations from the mean which means the data is not evenly scaled and contains significant outliers. To minimize outliers in the data, we chose to apply a data smoothing algorithm where each point is replaced with the average of the $n$ points on either side of it and itself. We see in Fig.~\ref{fig:scaled_phi_plots}(b) with $n=1$ that the data appears much more evenly scaled, but some points as still as far as 8 standard deviations from the mean. Likewise with $n=2$ in Fig.~\ref{fig:scaled_phi_plots}(c) there are points as far as 6 standard deviations away. Using $n=3$ results in the most even scaling overall with a the scale being approximately symmetric. While this data smoothing does result in some minor information loss, as long as $n$ remains relatively small the prominent signal topology should be retained. We chose to use $n=3$ for generating forecasting models and for use with TADA for the remainder of this section. 

\begin{figure}[htbp]
  \begin{minipage}[b]{0.5\linewidth}
    \centering
    \includegraphics[width=\textwidth]{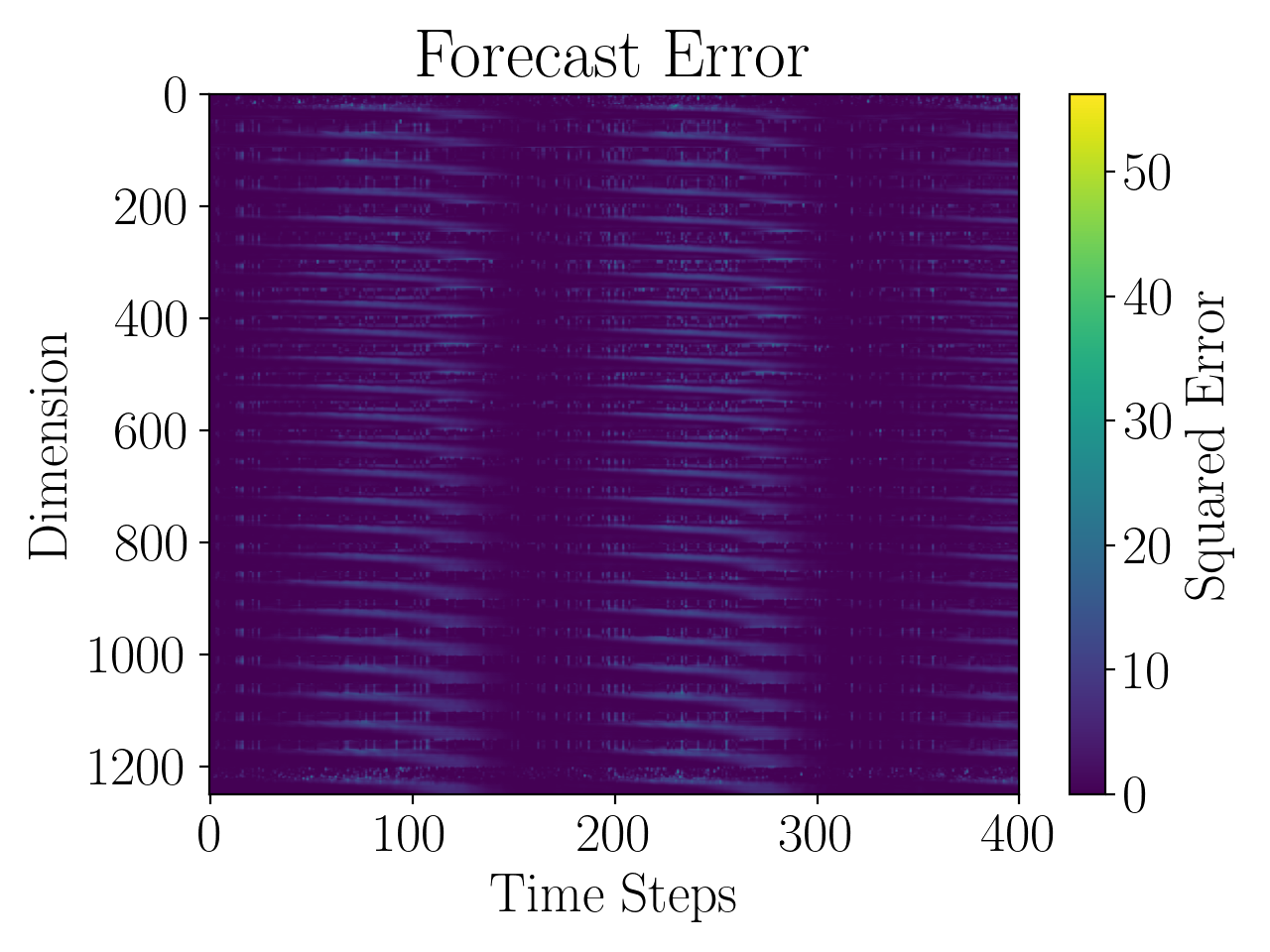}
    (a) Initial $\phi$ error
  \end{minipage}
  \begin{minipage}[b]{0.5\linewidth}
    \centering
    \includegraphics[width=\textwidth]{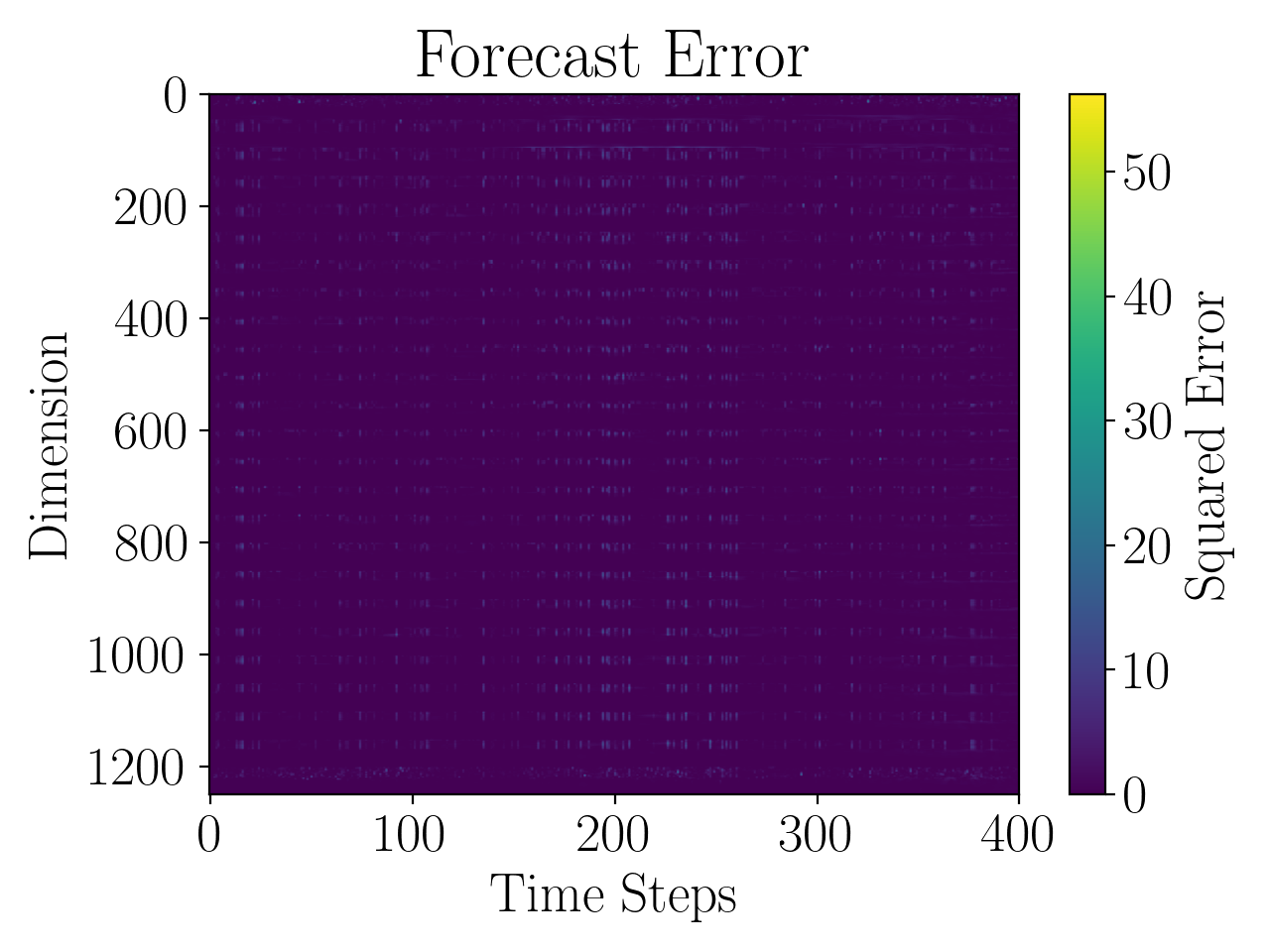}
    (b) Final $\phi$ error
  \end{minipage}
  \caption{Error plots for the electric potential before and after 200 TADA steps.}
  \label{fig:phi_error}
\end{figure}

\begin{figure}[htbp]
  \centering
  \includegraphics[width=0.75\textwidth]{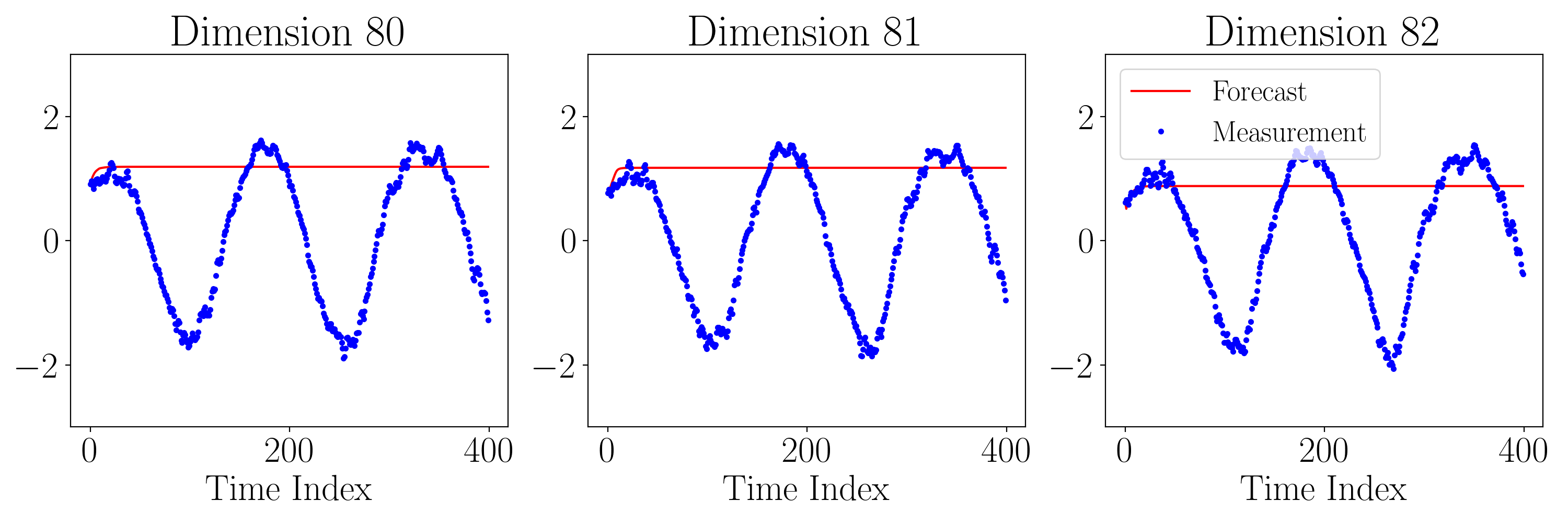}
  \caption{Initial $\phi$ forecast for dimensions 80-82.}
  \label{fig:phi_initial_forecast}
\end{figure}

The data smoothing also allows for more flexibility in the input model for TADA because the training data is cleaner and more accurately represents the structure in the signals. Therefore, we generated an LSTM model using the same parameters used for the electron temperature, but this time with only one epoch of optimization. The remainder of the forecasting and DA is handled by TADA. The initial forecast and error plots are shown in Figures~\ref{fig:phi_error}(a) and \ref{fig:phi_initial_forecast} where we see that the forecast for the unoptimized model is significantly different from the measurements and it does not capture any oscillations from the training data yet.  

\begin{figure}[htbp]
  \centering
  \includegraphics[width=0.75\textwidth]{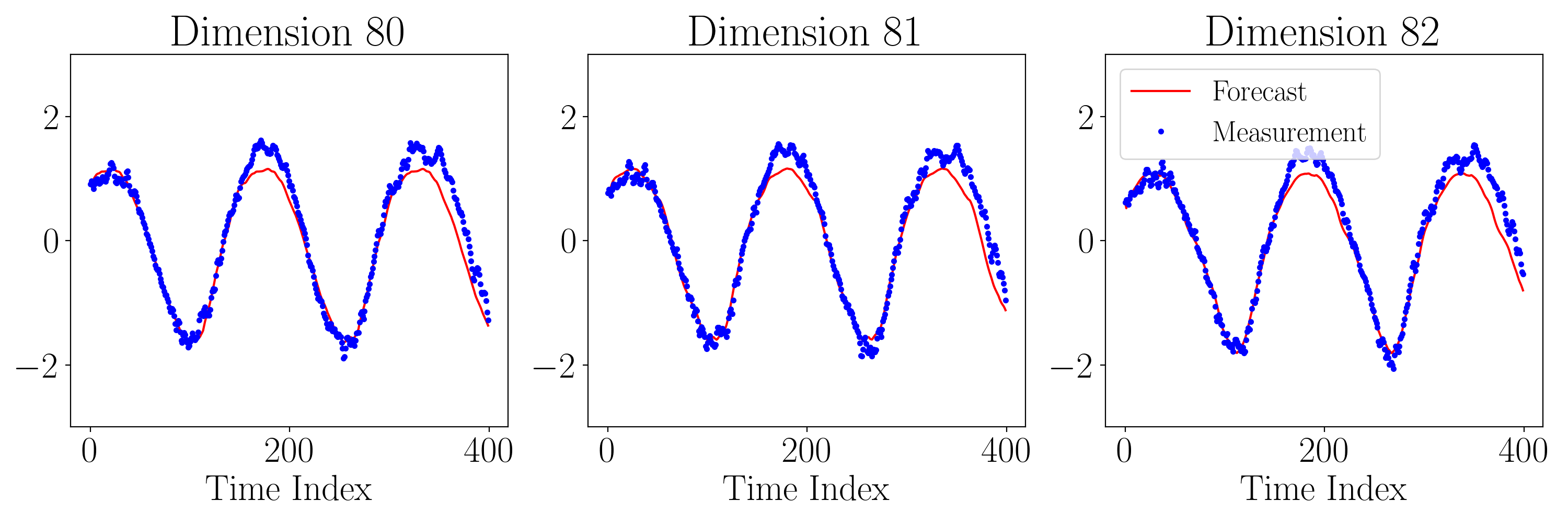}
  \caption{$\phi$ forecast for dimensions 80-82 after 200 TADA steps.}
  \label{fig:phi_200_forecast}
\end{figure}

TADA was applied to this model using a larger learning rate of $10^{-4}$ this time due to only one epoch of optimization occurring prior to using TADA. After 200 TADA steps, the resulting forecast error and forecast for dimensions 80--82 are shown in Figs.~\ref{fig:phi_error}(b) and \ref{fig:phi_200_forecast}.

We see that the forecast is significantly improved after applying TADA. The improvement compared to electron temperature is not as good due to the presence of significantly more noise in the $\phi$ field; however, TADA was still able to optimize the model such that it captures oscillations in the signal and provide a much more accurate forecast for this system. Note that the maximum error was initially over 50 and the final maximum error was approximately 27 with other points being significantly smaller.
\section{Conclusions}\label{sec:conclusion}
The TADA algorithm was successfully applied to high-fidelity, high-dimensional simulation data for an SPT-100 Hall Effect Thruster and many generalizations of the original algorithm were presented. We fit the LSTM network forecast function to the generalized equation for TADA and used this forecast function to generate data driven models for HETs. These models were then optimized using the full TADA cost function beginning with a forecast that is not close to the measurements and using persistence optimization to simultaneously learn from the training data and incoming measurements to provide an accurate forecast. Two HET field variables were tested with the algorithm, electron temperature and electric potential. It was found that if the data contains significant outliers after scaling that the forecast functions and TADA will not give accurate predictions. To fix this issue, a data smoothing algorithm was first applied to the data prior to scaling and this allowed TADA to accurately tune the model weights so the forecast accuracy improves.

\section{Acknowledgements}
This work is supported in part by Michigan State University and the National Science Foundation Research Traineeship Program (DGE-2152014) to Max Chumley. We would also like to acknowledge our collaborators at the Air Force Research Laboratory (AFRL) at Edwards Air Force Base. Specifically, we would like to thank David Bilyeu and Adrian Wong for generating and providing the data for this analysis.

\bibliographystyle{ieeetr}

\bibliography{Sections/bibliography}

\begin{thebibliography}{10}

\bibitem{Bapat2022}
A.~Bapat, P.~B. Salunkhe, and A.~V. Patil, ``Hall-effect thrusters for deep-space missions: A review,'' {\em IEEE Transactions on Plasma Science}, vol.~50, no.~2, pp.~189--202, 2022.

\bibitem{Hara2019}
K.~Hara, ``An overview of discharge plasma modeling for hall effect thrusters,'' {\em Plasma Sources Science and Technology}, vol.~28, no.~4, p.~044001, 2019.

\bibitem{chumley2024topological}
M.~M. Chumley and F.~A. Khasawneh, ``Topological approach for data assimilation,'' {\em arXiv preprint arXiv:2411.18627}, 2024.

\bibitem{Fife1998}
J.~M. Fife, {\em Hybrid-PIC modeling and electrostatic probe survey of Hall thrusters}.
\newblock PhD thesis, Massachusetts Institute of Technology, 1998.

\bibitem{Smith_2013}
B.~D. Smith, I.~D. Boyd, H.~Kamhawi, and W.~Huang, ``Hybrid-pic modeling of a high-voltage, high-specific-impulse hall thruster,'' in {\em 49th AIAA/ASME/SAE/ASEE Joint Propulsion Conference}, American Institute of Aeronautics and Astronautics, July 2013.

\bibitem{CohenSteiner2006}
D.~Cohen-Steiner, H.~Edelsbrunner, and J.~Harer, ``Stability of persistence diagrams,'' {\em Discrete {\&} Computational Geometry}, vol.~37, pp.~103--120, dec 2006.

\bibitem{Hatcher2002}
A.~Hatcher, {\em Algebraic Topology}.
\newblock Cambridge University Press, 2002.

\bibitem{Kaczynski2004}
T.~Kaczynski, K.~Mischaikow, and M.~Mrozek, {\em Computational Homology}.
\newblock Springer, Jan. 2004.

\bibitem{Ghrist2008}
R.~Ghrist, ``Barcodes: The persistent topology of data,'' {\em Builletin of the American Mathematical Society}, vol.~45, pp.~61--75, 2008.
\newblock Survey.

\bibitem{Carlsson2009}
G.~Carlsson, ``Topology and data,'' {\em Bulletin of the American Mathematical Society}, vol.~46, pp.~255--308, 1 2009.
\newblock Survey.

\bibitem{Edelsbrunner2010}
H.~Edelsbrunner and J.~Harer, {\em Computational Topology: An Introduction}.
\newblock Rhode Island: American Mathematical Society, 2010.

\bibitem{Mischaikow2013}
K.~Mischaikow and V.~Nanda, ``Morse theory for filtrations and efficient computation of persistent homology,'' {\em Discrete \& Computational Geometry}, vol.~50, no.~2, pp.~330--353, 2013.

\bibitem{oudot2017persistence}
S.~Y. Oudot, {\em Persistence theory: from quiver representations to data analysis}, vol.~209 of {\em AMS Mathematical Surveys and Monographs}.
\newblock Rhode Island: American Mathematical Soc., 2017.

\bibitem{Munch2017}
E.~Munch, ``A user's guide to topological data analysis,'' {\em Journal of Learning Analytics}, vol.~4, pp.~47--61, jul 2017.

\bibitem{Carriere2020}
M.~Carriere, F.~Chazal, M.~Glisse, Y.~Ike, and H.~Kannan, ``Optimizing persistent homology based functions,'' in {\em ICML 2021 - 38th International Conference on Machine Learning}, pp.~1294--1303, 2021.

\bibitem{Leygonie2021}
J.~Leygonie, S.~Oudot, and U.~Tillmann, ``A framework for differential calculus on persistence barcodes,'' {\em Foundations of Computational Mathematics}, vol.~22, pp.~1069--1131, 7 2021.

\bibitem{Gameiro2016}
M.~Gameiro, Y.~Hiraoka, and I.~Obayashi, ``Continuation of point clouds via persistence diagrams,'' {\em Physica D: Nonlinear Phenomena}, vol.~334, pp.~118--132, 11 2016.

\bibitem{Staudemeyer2019}
R.~C. Staudemeyer and E.~R. Morris, ``Understanding lstm -- a tutorial into long short-term memory recurrent neural networks,'' 2019.

\bibitem{geron2022hands}
A.~G{\'e}ron, {\em Hands-on machine learning with Scikit-Learn, Keras, and TensorFlow}.
\newblock " O'Reilly Media, Inc.", 2022.

\bibitem{zhang2003time}
G.~P. Zhang, ``Time series forecasting using a hybrid arima and neural network model,'' {\em Neurocomputing}, vol.~50, pp.~159--175, 2003.

\bibitem{mouraud2017innovative}
A.~Mouraud, ``Innovative time series forecasting: auto regressive moving average vs deep networks,'' {\em Entrepreneurship and Sustainability Issues}, vol.~4, no.~3, p.~282, 2017.

\bibitem{siami2018comparison}
S.~Siami-Namini, N.~Tavakoli, and A.~S. Namin, ``A comparison of arima and lstm in forecasting time series,'' in {\em 2018 17th IEEE international conference on machine learning and applications (ICMLA)}, pp.~1394--1401, Ieee, 2018.

\end{thebibliography}

\end{document}